\documentclass[preprint,12pt,authoryear]{elsarticle}

\usepackage{algorithm}  
\usepackage{algorithmicx}  
\usepackage{changepage}
\usepackage{hyperref}
\usepackage{multirow}
\usepackage[noend]{algpseudocode}
\usepackage{makecell}
\usepackage{amssymb}
\usepackage{amsmath}

\journal{}

\begin{document}

\begin{frontmatter}

\title{Hyperspectral Anomaly Detection with Self-Supervised Anomaly Prior}

\author[label1]{Yidan Liu}
\author[label1]{Weiying Xie \corref{cor1}}
\author[label1]{Kai Jiang}
\author[label1]{Jiaqing Zhang}
\author[label1]{Yunsong Li}
\author[label2]{Leyuan Fang}

\affiliation[label1]{organization={State Key Laboratory of Integrated Services Networks, Xidian University},
            city={Xi'an},
            postcode={710071},
            country={China}}

\affiliation[label2]{organization={College of Electrical and Information Engineering, Hunan University},
            city={Changsha},
            postcode={410082},
            country={China}}

\cortext[cor1]{Corresponding author: wyxie@xidian.edu.cn (Weiying Xie)}

\begin{abstract}
			The majority of existing hyperspectral anomaly detection (HAD) methods use the low-rank representation (LRR) model to separate the background and anomaly components, where the anomaly component is optimized by handcrafted sparse priors (e.g., $\ell_{2,1}$-norm). However, this may not be ideal since they overlook the spatial structure present in anomalies and make the detection result largely dependent on manually set sparsity. To tackle these problems, we redefine the optimization criterion for the anomaly component in the LRR model with a self-supervised network called self-supervised anomaly prior (SAP). This prior is obtained by the pretext task of self-supervised learning, which is customized to learn the characteristics of hyperspectral anomalies. Specifically, this pretext task is a classification task to distinguish the original hyperspectral image (HSI) and the pseudo-anomaly HSI, where the pseudo-anomaly is generated from the original HSI and designed as a prism with arbitrary polygon bases and arbitrary spectral bands. In addition, a dual-purified strategy is proposed to provide a more refined background representation with an enriched background dictionary, facilitating the separation of anomalies from complex backgrounds. Extensive experiments on various hyperspectral datasets demonstrate that the proposed SAP offers a more accurate and interpretable solution than other advanced HAD methods.   
\end{abstract}


\begin{keyword}
  Self-supervised learning (SSL) \sep Anomaly detection \sep Deep prior \sep Low-rank representation (LRR) \sep Hyperspectral image (HSI).
\end{keyword}

\end{frontmatter}



\section{Introduction}
\label{sec1}
Hyperspectral image (HSI) has become a favorable tool for deep space exploration and earth observation due to its fine-grained spectral vectors that can reflect the physical properties of materials \citep{intro21,NN1,NN2}. Hyperspectral anomaly detection (HAD), one of the research hotspots in HSI processing, aims to identify and locate unusual targets in the space or spectral domain without a reference spectrum, where the 'unusual targets' are also called anomalies \citep{HAD1,HAD2}.

Generally, the occurrence probability of anomalous pixels is relatively low, and their correlation is not as strong as that among background pixels. In view of this, the low-rank representation (LRR)-based methods have been developed \citep{LRASR}. They typically decompose HSI into background and anomaly components, and use the nuclear norm ${\left\|  \cdot  \right\|_*}$ and $\ell_{2,1}$-norm ${\left\|  \cdot  \right\|_{2,1}}$ as optimization criteria for these components to establish a mathematical optimization problem
\begin{equation}
\begin{gathered}
\min _{\mathbf{E}, \mathbf{A}}\|\mathbf{E}\|_*+\beta\|\mathbf{A}\|_{2,1} \\
\text { s.t. } \mathbf{H}=\mathbf{D E}+\mathbf{A}.
\end{gathered}
\label{LRR}
\end{equation}
Here, ${\bf{H}}$ denotes the observed HSI, ${\bf{A}}$ represents the anomaly component, and ${\bf{D E}}$ represents the background component. The parameter $\beta $, as the sparsity of anomalies, is manually adjustable. In this way, Eq. (\ref{LRR}) can be efficiently solved by the alternating direction multiplier method (ADMM) \citep{ADMM}, enabling the anomaly component to be separated.

Despite the theoretical appeal of LRR-based methods, they still suffer from some problems, resulting in limited performance. Firstly, existing LRR-based methods mainly use handcrafted sparse priors (e.g., $\ell_{2,1}$-norm) to optimize the anomaly component, making the detection result largely dependent on the manually set sparsity, reducing the flexibility of practical applications. Secondly, these priors primarily concentrates on the sparsity of spectral vectors, lacking the spatial information of anomalies. For example, aircraft usually appear as structured areas rather than isolated points. Therefore, using a uniform handcrafted prior to optimize the anomaly component may not be ideal.

Fortunately, the plug-and-play strategy would be a potential way to address this issue. Its main idea is to utilize well-performed priors obtained from available techniques to solve the sub-optimization problem separated from the target model, rather than designing new priors. With this strategy, enormous progress has been made in image reconstruction \citep{PAP_HSI1,PAP_otherfield3,PAP_HSI2}. Recently, \citep{PAP} utilizes a denoising convolutional neural network (DeCNN) as the low-rank prior for the background, introducing the plug-and-play strategy into HAD for the first time.

Inspired by these works, an idea naturally comes into our minds: Is it possible to use a deep neural network (DNN) as the anomaly prior to eliminate the limitations of handcrafted priors? To this end, DNN needs to learn the sparsity and spatial characteristics of anomalies. The former is an inherent statistical property of the anomaly component, while the latter is a flaw in the regular LRR model. Over past decades, there have emerged a series of DNN-based HAD methods, and some of them already meet the above two requirements \citep{RGAE,spatialHAD}. However, these methods always design the network with shallow layers, which limits the network's generalization ability. Therefore, it is not suitable to use these networks as anomaly prior, and it is necessary to design a network specifically for hyperspectral anomaly prior.

Given the lack of labels and training samples in HAD, we adopt self-supervised learning to train the network. In self-supervised learning, the pretext task can "automatically" generate supervisory signals for training \citep{SSL}, which are derived from the input data itself rather than manually annotated labels provided alongside the input data. Thus, training the network in a self-supervised manner only requires original HSIs, avoiding the problem of lacking trainable labels in the HAD. In addition, we design the pretext task as distinguishing the original image and the image with pseudo-anomaly, in which the generation of pseudo-anomaly allows us to use hyperspectral classification datasets without anomalies for network training. These classification datasets typically have a large number of images, and they can be further increased with data augmentation methods, effectively addressing the problem that the HAD dataset is too small to train the network.

In summary, we propose a novel LRR model with \textbf{S}elf-supervised \textbf{A}nomaly \textbf{P}riors \textbf{(SAP)} for HAD. To obtain a prior that can generalize various anomalies, we first customize the pretext task of self-supervised learning by comprehensively considering both the sparsity and spatial characteristics of hyperspectral anomalies. Specifically, this pretext task is a classification task to distinguish the original HSI and the HSI with pseudo-anomaly, where the pseudo-anomaly image is generated from the original image. Then, the body part of the network trained by the pretext task, as the anomaly prior, is plugged into the LRR model to solve the anomaly sub-problem, which corresponds to the target task of self-supervised learning. Furthermore, to enhance anomaly-background separability, we design a dual-purified strategy to provide a more refined background representation with an enriched background dictionary, enabling the anomalies to be more easily separated from complex backgrounds. Overall, our work makes three main contributions:

\begin{enumerate}
	\item SAP is the first study to obtain the anomaly prior by deep learning in the HAD field, which not only comprehensively considers the spatial and spectral characteristics of anomalies, but is also independent of manually set parameters.
    \item To obtain a prior that can generalize various anomalies in the lack of labels and limited samples, a customized pretext task is designed as distinguishing the original HSI and the pseudo-anomaly HSI generated from the original HSI.
	\item To obtain an enriched background dictionary without anomaly contamination, a dual-purified strategy for dictionary construction is proposed. Extensive experiments on real hyperspectral datasets from different sensors demonstrate that SAP provides more accurate background representation and state-of-the-art results.
\end{enumerate} \par

The remainder of this paper is organized as follows: Section \ref{sec2} reviews existing works on LRR-based and DNN-based HAD methods. Section \ref{sec3} details the proposed SAP, and Section \ref{sec4} presents the experimental results and discussion. Finally, Section \ref{sec5} reports the conclusions of our study.

\section{Related Work}
\label{sec2}
\subsection{LRR-based HAD}
\label{subsec21}
Earlier LRR-based HAD methods have not introduced the background dictionary, referred to as the low-rank and sparse matrix decomposition model (LSDM). Later, various LSDM-based methods have developed. For example, \citep{MOG} replaces the sparse matrix with the mixture of Gaussian (MoG), and \citep{OSP} performs go decomposition (GoDec) in orthogonal subspace-projections (OSP) to determine two matrixex. As an improvement of LSDM, LRR model was first applied in the HAD field by \citep{LRASR}, where the low-rank constraint and sparsity constraint are simultaneously imposed on the representation coefficients of the dictionary (LRASR). Subsequently, numerous variants were proposed. \citep{GTVLRR} incorporates the spatial geometrical structure into the LRR model, \citep{low-dimensional} locates anomalous pixels in a low-dimensional feature space created by data-driven projection, and \citep{tensorTC} uses tensor representation to maintain the intrinsic 3-D structure of HSI, avoiding destruction from 2-D conversion.

Some works on combining deep learning and LRR models have been carried out, like \citep{AAAI,WSLRR}, but they essentially construct a comprehensive dictionary through deep learning without making DNN a part of the LRR model.

\subsection{DNN-based HAD}
\label{subsec22}
The boom of DNN in natural image processing stimulates researchers to apply deep learning to the HAD task. Considering the difficulty of obtaining labels, existing DNN-based HAD methods always train the network in an unsupervised manner. For example, \citep{SCAAE} and \citep{DualAE} employ autoencoder (AE) as feature extraction network to provide more discriminative features for the subsequent anomaly detector. Considering such separated detection methods may yield a suboptimal solution, \citep{E2E} designs a unified framework in which the estimation network, as an anomaly detector, is connected after AE and optimized in an end-to-end manner. \citep{memoryAE} proposes an AE with memory modules to prevent detection accuracy from being reduced by reconstructing anomalies. Additionally, a number of studies in designing more feasible DNN-based HAD methods have been developed. \citep{DFAE} addresses the problem of limited generalization of DNN-based HAD methds by detecting anomalies in the image frequency domain. \citep{new_generalization} achieves generalization on a large-scale dataset by learning an anomaly enhancement network through a data augmentation strategy called random mask. Nonetheless, unsupervised HAD performance is still restricted by lacking the prior information of anomalies.

To catch up with the performance of supervised learning, \citep{semisupervised1} and \citep{semisupervised2} propose semi-supervised HAD methods. Besides, \citep{weaklyAD} introduces the concept of weakly supervised learning into HAD, opening up more flexible ways to study HAD. However, these methods have an inherent problem of not obtaining pure and comprehensive background labels, which may make them sink into suboptimal solutions. Fortunately, self-supervised learning can effectively avoid this problem. As a branch of unsupervised learning, self-supervised learning does not require any human annotations but provides promising results that are comparable to supervised learning \citep{SSL}. In general, self-supervised learning consists of a pretext task and a target task. The former assists the network in learning the intrinsic features of input data, and the latter contains the first several layers of the well-trained network as a feature extractor or a pre-trained model. It is worth noting that pseudo labels are generated from input data at the beginning of the pretext task, which can be viewed as matching the concept of supervised learning to approach its superior performance. Therefore, self-supervised learning has been applied in training HAD networks, specifically for background modeling \citep{SSLHAD1,SSLHAD2}.

\begin{figure*}[t]
\centering
\includegraphics[width=\textwidth]{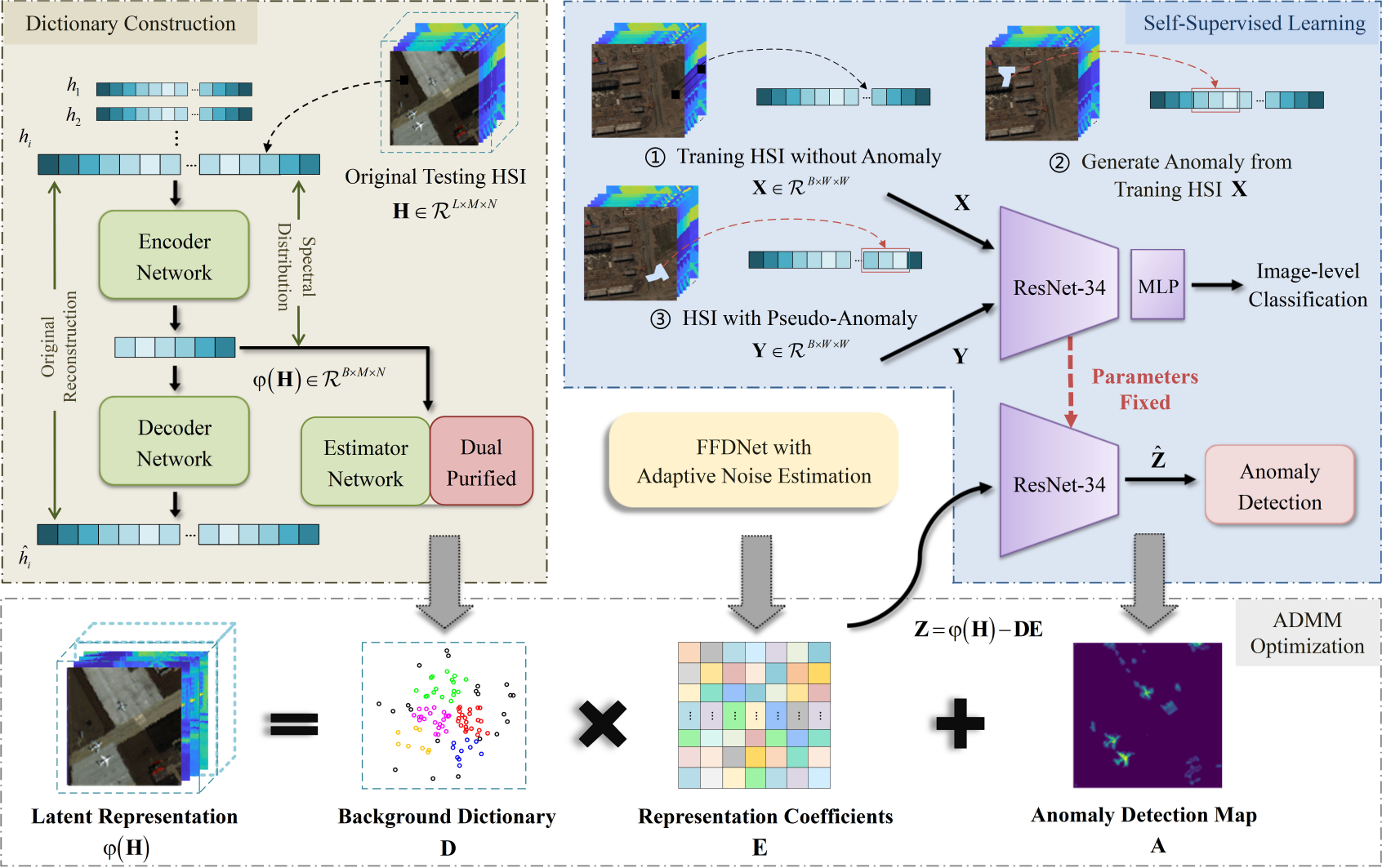}
\caption{Schematic of the proposed SAP method.}
\label{framefork}
\end{figure*}

\section{Proposed Method}
\label{sec3}
In this paper, we propose a self-supervised LRR for HAD, aiming to replace the handcrafted sparse prior with self-supervised learned DNN to solve the anomaly component. The objective function of our proposed SAP is formulated as 
\begin{align}
\begin{array}{c}
\mathop {\min }\limits_{{\bf{E}},{\bf{A}},{\bf{J}}} \frac{1}{2}\left\| {\varphi\left( {\bf{H}} \right){\bf{ - DE - A}}} \right\|_F^2 + \psi \left( {\bf{J}} \right) + \gamma ({\bf{A}})\\[1mm]
s.t.{\rm{\quad}}{\bf{E = J}}
\end{array}
\label{objective-function}
\end{align}
where $\gamma ( \cdot )$ denotes the anomaly prior obtained by self-supervised learning, and ${\bf{A}} \in {{\cal R}^{B \times MN}}$ stands for the anomaly matrix with $MN$ pixels and $B$ spectral bands. $\varphi \left( {\bf{H}} \right) \in {{\cal R}^{B \times MN}}$ is a latent HSI extracted from the unfolded ${\bf{H}} \in {{\cal R}^{L \times M \times N}}$ to keep the dimensionality consistent with the anomaly prior $\gamma ( \cdot )$. ${\bf{D}} \in {{\cal R}^{B \times nb}}$ denotes the dictionary matrix and consists of $nb$ background pixels selected from $\varphi \left( {\bf{H}} \right)$ as dictionary atoms. ${\bf{E}} \in {{\cal R}^{nb \times MN}}$ is the representation coefficient of ${\bf{D}}$, ${\bf{J}}$ is an auxiliary variable for ${\bf{E}}$ to decouple the data item and the prior items, and $\psi \left(  \cdot  \right)$ represents a denoising network with adaptive noise estimation to replace the nuclear norm as the low-rank prior.

Thus, the proposed method primarily consists of two phases: firstly, utilizing self-supervised learning to train the network to get the anomaly prior, and secondly, solving the LRR model (\ref{objective-function}) with the well-trained network. In other words, the well-trained network is the concrete form of the anomaly prior, and it also called self-supervised prior. Notably, the construction of the dictionary ${\bf{D}}$ and latent HSI $\varphi \left( {\bf{H}} \right)$ is separate from the above process since they are constants in the optimization of the LRR model, and we will introduce it in Section \ref{subsec33}. The schematic of our proposed SAP is presented in Fig. \ref{framefork}

\begin{figure}[t]
\centering
\includegraphics[width=\textwidth]{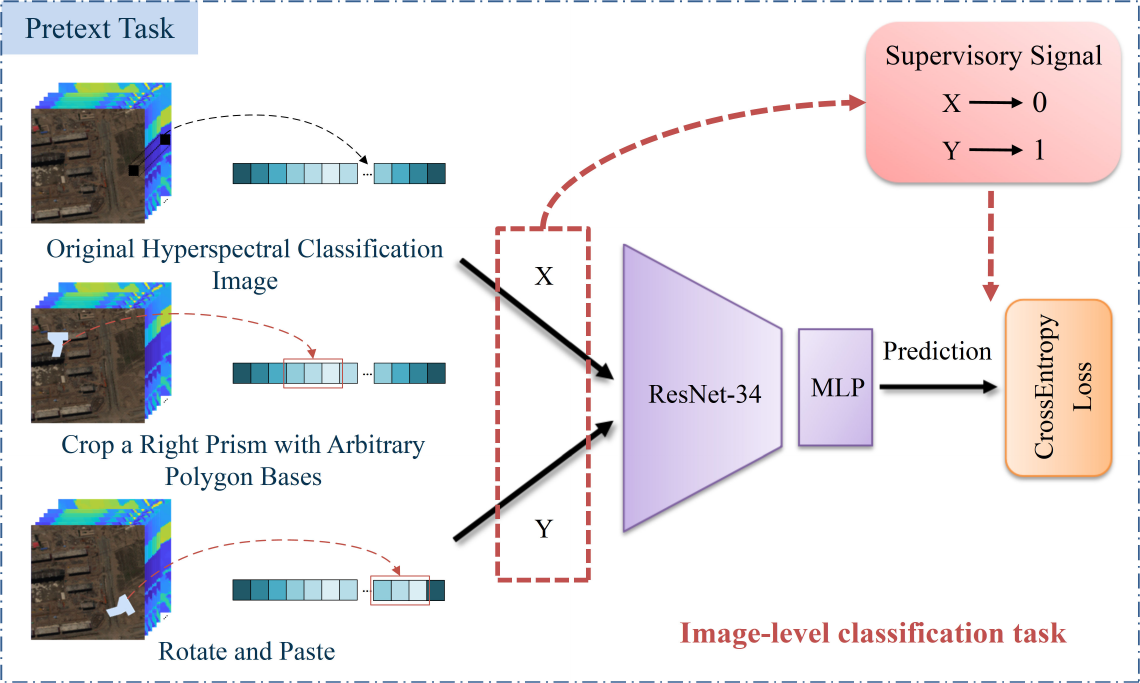}
\caption{The flowchart of the pretext task for self-supervised learning.}
\label{pretext_task}
\end{figure}

\subsection{Learning a Self-Supervised Prior}
\label{subsec31}
As mentioned earlier, self-supervised learning is composed of two parts: the pretext task and the target task. In this paper, the pretext task aims to train the network to capture the data characteristics of hyperspectral anomalies, while the target task is to solve the anomaly sub-problem, i.e., $\gamma ({\bf{A}})$ in Eq. (\ref{objective-function}).

As shown in Fig. \ref{pretext_task}, our pretext task is designed as an image-level classification task, with ResNet34 as the backbone network and connected to a multi-layer perceptron (MLP) as the detection head. The inputs to the network are the original hyperspectral classification image ${\bf{X}} \in {{\cal R}^{B \times W \times W}}$ and the HSI with pseudo-anomaly ${\bf{Y}} \in {{\cal R}^{B \times W \times W}}$, where ${\bf{Y}}$ is generated from ${\bf{X}}$. Specifically, the generation of ${\bf{Y}}$ can be viewed as three steps:
\begin{enumerate}
\setlength{\parsep}{2pt}
\setlength{\parskip}{2pt}
    \item First, crop a prism from any spatial and spectral position of ${\bf{X}}$, which has arbitrary polygon bases and an arbitrary number of bands.
    \item Then, rotate the cropped prism at any angle.
    \item Finally, paste the rotated prism onto any position and any band of ${\bf{X}}$.
\end{enumerate}

Compared with general data augmentation methods, our anomaly generation approach takes the sparsity of anomalies into account at first, that is, there is only one pseudo anomaly in each image, occupying a small area. Secondly, the spatial morphology of anomalies is preserved by the design of the polygonal structure. More importantly, the size, shape, angle, and position of each pseudo-anomaly are all arbitrary and uncertain. This multi-dimensional randomness ensures the irregularity and unexpectedness of the anomalies. Furthermore, given that some anomalies only differ from the background in certain spectra, the pseudo-anomaly is specially designed with an "arbitrary number of bands." 

While generating ${\bf{Y}}$, the corresponding supervisory signals are also generated for two inputs, that is, the signal 0 for ${\bf{X}}$ and signal 1 for ${\bf{Y}}$. These signals participate as labels in network training, and the loss function is defined as the cross entropy of the difference between the network's predictions and these labels:
\begin{align}
Loss =  - \frac{1}{N}\sum\limits_{i = 1}^N {\left( {p\left( {{x_i}} \right)\log q\left( {{x_i}} \right) + p\left( {{y_i}} \right)\log q\left( {{y_i}} \right)} \right)}
\label{loss}
\end{align}
where ${p\left( {{x_i}} \right)}$ and ${p\left( {{y_i}} \right)}$ represent the one-hot vectors of supervisory signals 0 and 1 respectively. ${q\left(  \cdot  \right)}$ denotes the softmax result of network's predictions, and $N$ stands for the total number of HSIs participating in the training process. After completing the pretext task, the parameters of ResNet34 are fixed, and it can extract the data that similar to the learned hyperspectral anomalies in the subsequent target task.

\subsection{LRR Model with Self-Supervised Prior}
\label{subsec32}
Mathematically, our model (\ref{objective-function}) is solved as follows: First, convert to an unconstrained optimization problem
\begin{align}
\mathop {\min }\limits_{{\bf{E}},{\bf{A}},{\bf{J}},{\bf{L}}} \frac{1}{2}{\left\| {\varphi \left( {\bf{H}} \right){\bf{ - DE - A}}} \right\|_F} + \psi \left( {\bf{J}} \right) + \gamma ({\bf{A}}) \nonumber\\
 + \frac{\alpha }{2}\left\| {{\bf{J - E}} + \frac{{\bf{L}}}{\alpha }} \right\|_F^2 - \frac{1}{{2\alpha }}\left\| {\bf{L}} \right\|_F^2
\label{ALM}
\end{align}
where ${\bf{L}}$ is the Lagrange multiplier and $\alpha > 0$ is the penalty parameter. Next, split Eq. (\ref{ALM}) by ADMM and update each variable iteratively.

\subsubsection{${\bf{E}}$ step}
With ${\bf{A}}$, ${\bf{J}}$, ${\bf{L}}$ fixed, ${\bf{E}}$ is updated by
\begin{align}
\mathop {\min }\limits_{\bf{E}} \frac{1}{2}\left\| {\varphi \left( {\bf{H}} \right) - {\bf{DE}} - {\bf{A}}} \right\|_{_F}^2 + \frac{\alpha }{2}\left\| {{\bf{J}} - {\bf{E}} + \frac{{\bf{L}}}{\alpha }} \right\|_{_F}^2
\label{S}
\end{align}
As a strongly convex function, the minimum ${\bf{E}}$ can be easily found to be
\begin{align}
{\bf{E}} = {\left( {{{\bf{D}}^T}{\bf{D}} + \alpha {\bf{I}}} \right)^{ - 1}}\left[ {{{\bf{D}}^T}\left( {\varphi \left( {\bf{H}} \right) - {\bf{A}}} \right) + \alpha {\bf{J}} + {\bf{L}}} \right]
\label{S_res}
\end{align}

\subsubsection{${\bf{J}}$ step}
With ${\bf{E}}$, ${\bf{A}}$, ${\bf{L}}$ fixed, ${\bf{J}}$ is updated by
\begin{align}
\mathop {\min }\limits_{\bf{J}} \psi \left( {\bf{J}} \right) + \frac{\alpha }{2}\left\| {{\bf{J}} - {\bf{E}} + \frac{{\bf{L}}}{\alpha }} \right\|_{_F}^2
\label{W}
\end{align}
According to \citep{PAP}, the optimization of Eq. (\ref{W}) can be solved with a CNN-based denoiser FFDNet \citep{ffdnet}, whose input noisy data is $\left( {{\bf{E}} - {{\bf{L}} \mathord{\left/
 {\vphantom {{\bf{L}} \alpha }} \right.
 \kern-\nulldelimiterspace} \alpha }} \right)$. However, FFDNet requires input of noise distribution parameters during processing, which are manually set in \citep{PAP} and need to be reset for different HSIs, limiting its flexibility.

To address this problem, we adopt a statistics-based approach \citep{noise} to estimate the noise level. Based on the observation that most noise-free images lie in low-dimensional subspaces, this approach effectively converts the noise estimation problem into a search for redundant dimensions of the observed data by calculating the covariance matrix and eigenvalue, thereby adaptively estimating the noise.

\begin{algorithm}[t]
	\caption{The Target task for Self-Supervised Learning}
	{\bf{Input:}} ${\bf{Z}} = \varphi \left( {\bf{H}} \right) - {\bf{DE}}$
	\par \hspace*{0.01in}
        {\bf{Step:}}
    \hspace*{0.3in}
	\label{alg1}
	\begin{algorithmic}[1]
		\State Split ${\bf{Z}}$ into small cubes;
		\State Extract feature vectors ${\bf{\hat Z}}$ through the network learned from the pretext task;
		\State Calculate the anomaly score ${{{\rm{\hat A}}}_i}$ of each small cube using Mahalanobis distance;
		\State Propagate the anomaly score by Gaussian smoothing, and obtain the initial detection map ${{\rm{\hat A}}}$;
		\State Perform adaptive threshold segmentation on ${{\rm{\hat A}}}$ to generate the guided map ${{{\rm{\hat A}}}_G}$;
        \State Update ${\bf{A}}$ by ${\bf{A}} = {{{\rm{\hat A}}}_G} \cdot {\bf{Z}}$.
	\end{algorithmic}
	{\bf{Output:}} ${\bf{A}}$
\label{E_updating} 
\end{algorithm}

\subsubsection{${\bf{A}}$ step}
With ${\bf{E}}$, ${\bf{J}}$, ${\bf{L}}$ fixed, ${\bf{A}}$ is updated by
\begin{align}
\mathop {\min }\limits_{\bf{A}} \gamma \left( {\bf{A}} \right) + \frac{1}{2}\left\| {\varphi \left( {\bf{H}} \right) - {\bf{DE}} - {\bf{A}}} \right\|_{_F}^2
\label{E}
\end{align}
Here, $\gamma ({\bf{A}})$ indicates that ${\bf{A}}$ is complied with the $\gamma ( \cdot )$ that contains the characteristics of hyperspectral anomalies, and $\left\| {\varphi \left( {\bf{H}} \right) - {\bf{DE}} - {\bf{A}}} \right\|_{_F}^2$ is the fidelity term. Thus, the optimization of Eq. (\ref{E}) is solved by inputting ${\bf{Z}} = \varphi \left( {\bf{H}} \right) - {\bf{DE}}$ into $\gamma ( \cdot )$. This process is regarded as the target task of self-supervised learning and the specific operation steps is summarized in Algorithm \ref{E_updating}.

Notably, the output of the ResNet34 learned by the pretext task is a one-dimensional vector \citep{resnet}. This indicates that the well-trained ResNet34 acts as a feature extractor that can capture anomaly features but cannot directly produce anomaly scores, let alone the detection map. Hence, it is not feasible to input the whole image ${\bf{Z}}$ into the network.

In light of this, ${\bf{Z}} \in {{\cal R}^{B \times M \times N}}$ is divided into $K$ overlapped small cubes before entering the well-trained ResNet34, and the anomaly scores for their feature vectors ${\bf{\hat Z}} \in {{\cal R}^{K \times F}}$ output from the network are computed by the Mahalanobis distance
\begin{align}
{{{\rm{\hat E}}}_i} = {\left( {{{{\bf{\hat Z}}}_i} - \mu } \right)^T}{{\bf{\Gamma }}^{ - 1}}\left( {{{{\bf{\hat Z}}}_i} - \mu } \right)
\label{RX}    
\end{align}
where ${{{\rm{\hat E}}}_i}$ denotes the anomaly score of the $i$-th small cube. ${{{\bf{\hat Z}}}_i} \in {{\cal R}^{1 \times F}}$ represents the feature vector corresponding to the $i$-th small cube, with the size of $F$ (i.e., 512 in ResNet34). $\mu  \in {{\cal R}^{1 \times F}}$ and ${\bf{\Gamma }} \in {{\cal R}^{F \times F}}$ are the mean vector and covariance matrix of ${\bf{\hat Z}}$, respectively. 

Since only one score can be calculated for each small cube, it is necessary to propagate the anomaly score to all the pixels occupied by the corresponding small cube through Gaussian smoothing \citep{deconvolution}. Essentially, the Gaussian smoothing operation follows the spirit of the deconvolution operation in deep learning, which aims to map the feature maps obtained by DNN back to the input image. The difference in our SAP is that a convolution kernel with fixed parameters is used instead of a trainable deconvolution layer.

After Gaussian smoothing, we get an initial detection map ${\rm{\hat A}} \in {{\cal R}^{M \times N}}$ with the same spatial size as ${\bf{Z}}$. Considering that multiple iterative updates may enhance the noise, the adaptive threshold segmentation is performed on ${{\rm{\hat A}}}$ to generate a guided map ${{{\rm{\hat A}}}_G} \in {{\cal R}^{M \times N}}$, which not only preserves the anomaly detection results but also prevents the accumulation of noise. As a result, the updated ${\bf{A}}$ is obtained by multiplying the guided map ${{{\rm{\hat A}}}_G}$ and the input ${\bf{Z}}$.

\subsubsection{${\bf{L}}$ step}
With ${\bf{E}}$, ${\bf{A}}$, ${\bf{J}}$ fixed, ${\bf{L}}$ is updated by
\begin{align}
{\bf{L}} = {\bf{L}} + \alpha \left( {{\bf{J}} - {\bf{E}}} \right)
\label{L}
\end{align}

\begin{figure}[t]
\centering
\includegraphics[width=\textwidth]{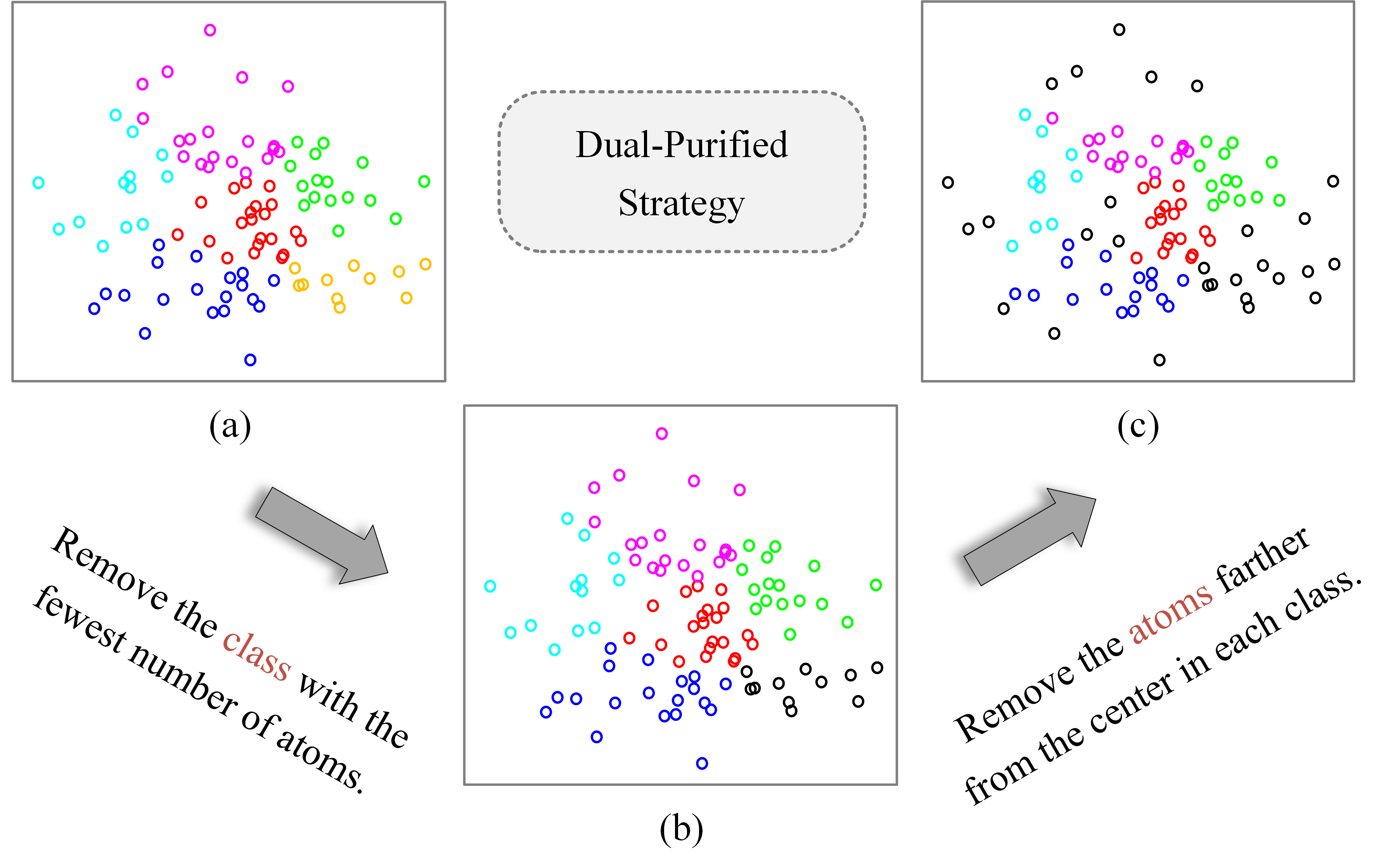}
\caption{Schematic of the proposed dual-purified strategy. (a) The output of estimator network. (b) and (c) are the purified results. The atoms farther from the class center indicate a lower probability of occurrence.}
\label{dictionary}
\end{figure}

\subsection{Construction of Dictionary and Latent HSI}
\label{subsec33}
When solving the LRR model, the observed HSI and background dictionary are fixed matrices, which means they need to be determined before optimization. However, the hyperspectral classification dataset we used for the pretext task (i.e., ${\bf{X}}$ in Section \ref{subsec31}) has only 48 spectral bands, but the HAD dataset ${\bf{H}}$ typically has around 200 bands. Therefore, it is necessary to reduce the dimensionality of the original HAD dataset to ensure consistency with the feature dimensions learned by self-supervised learning. Fortunately, \citep{AAAI} designs a low-rank embedded network (LREN) that can achieve both dimensionality reduction and dictionary construction, which aligns with the problem we are trying to solve. As shown in Fig. \ref{framefork}, LREN is composed of an AE network for extracting latent representation and an estimator network for spectral classification, which are jointly trained in an end-to-end manner.

Nonetheless, the size of the dictionary in \citep{AAAI} is too small to provide a good representation of the background since it only selects the center spectrum of each class and the spectral sample farthest from the class center as dictionary atoms. Therefore, we propose a dual-purified strategy to construct our background dictionary, enriching the dictionary atoms without anomaly contamination. Based on the fact that anomalies always occur in small proportions, the class with the fewest number of spectral samples is initially eliminated, followed by the removal of those samples with low probability in each class, making the remaining spectral samples the dictionary atoms. As illustrated in Fig. \ref{dictionary}, with the dual-purified strategy, the output of estimator network turns those suspicious anomaly atoms into black and composes the retained colored atoms as the dictionary in our proposed SAP.

Furthermore, considering the inherent spectral structure of HSI, we add a spectral distribution constraint (SDC) \citep{SDEN} to the loss function of LREN. This constraint is based on the manifold assumption, intending to keep the consistency of spectral distributions between the latent HSI $\varphi \left( {\bf{H}} \right)$ and the original image ${\bf{H}}$. More specifically, it is achieved by minimizing the spectral angular distance (SAD) between each spectral sample and the clustering center spectrum in ${\bf{H}}$ and $\varphi \left( {\bf{H}} \right)$, which is mathematically expressed as
\begin{equation}
\begin{gathered}
{L_{SDC}} = \frac{1}{C}\sum\limits_{i = 1}^N {{\rm{SAD}}\left( {{h_i},{h_c}} \right)}  - \frac{1}{C}\sum\limits_{i = 1}^N {{\rm{SAD}}\left( {\varphi \left( {{h_i}} \right),\varphi \left( {{h_c}} \right)} \right)} \\[1mm]
{\rm{SAD}}\left( {{h_i},{h_c}} \right) = \arccos \frac{{h_i^{\rm{T}}{h_c}}}{{{{\left\| {{h_i}} \right\|}_2}{{\left\| {{h_c}} \right\|}_2}}}
\end{gathered}
\label{SAD}
\end{equation}
Here, $N$ is the total number of spectral samples. ${{h_i}}$ and ${{h_c}}$ respectively represent the $i$-th spectral sample and the spectral sample with the minimum SAD to the spectral clustering center calculated by K-means \citep{K-means}, where $c \in N$. $C$ stands for the number of spectral clusters, which is set to 2 as \citep{SDEN}, indicating that all spectral samples are divided into two clusters: background and anomaly.

\begin{algorithm}[t]
	\caption{The proposed SAP for HAD}
	{\bf{Input:}} original HAD dataset ${\bf{H}}$, training HSI ${\bf{X}}$;
        \par
        {\bf{Initialize:}} $C = 2$, $\lambda  = {10^{ - 3}}$, $\alpha  = 1$, ${\bf{E}} = {\bf{A}} = {\bf{J}} = {\bf{L}} = {\bf{0}}$;
	\par
        {\bf{Step:}}
	\begin{algorithmic}[1]
		\State Generate the HSI with pseudo-anomaly ${\bf{Y}}$; 
		\State Obtain the well-trained ResNet34 from ${\bf{X}}$ and ${\bf{Y}}$ according to Eq. (\ref{loss});
		\State Construct $\varphi \left( {\bf{H}} \right)$ and ${\bf{D}}$ from ${\bf{H}}$ according to Eq. (\ref{loss_dict}) and dual-purified strategy;
		\State Optimize the problem in Eq. (\ref{objective-function}) with ADMM and plug-and-play strategy
            \Statex \textbf{while} no converged \textbf{do}
            \begin{itemize}
                \item Update ${\bf{E}}$ via Eq. (\ref{S_res}); \par
                \item Update ${\bf{J}}$ via FFDNet with adaptive noise estimation; \par
                \item Update ${\bf{A}}$ via Algorithm \ref{E_updating}; \par
                \item Update ${\bf{L}}$ via Eq. (\ref{L}); \par
            \end{itemize}
            \Statex \textbf{end while}
	\end{algorithmic}
	{\bf{Output:}} ${\bf{A}}$
\label{SAP} 
\end{algorithm}

As a result, the network for constructing the dictionary and latent HSI in SAP is trained by minimizing
\begin{align}
{L_{SAP}} = {L_{LREN}} + \lambda  \cdot {L_{SDC}}
\label{loss_dict}
\end{align}
where $\lambda $ is a hyperparameter determined experimentally. Other hyperparameters are consistent with those in \citep{AAAI}, except for the hidden nodes of the AE network, which is set to 48 to match the dimension of the self-supervised anomaly prior. The entire algorithm of SAP for HAD is summarized in Algorithm \ref{SAP}.

\section{Experiments}
\label{sec4}
\subsection{Data Description}
\label{subsec41}
\textbf{The traning dataset} used for self-supervised learning (see Section \ref{subsec31}) comes from \citep{traningHSI}. This dataset contains 1,385 HSIs, and each image has 256$\times$256 pixels with 48 spectral bands ranging from 380 to 1050 nm. Before being input into the network, they are cut into a size of 64$\times$64 pixels. Considering the scene content and the signal-to-noise ratios, we selected 873 images from the entire dataset, and divided them at 4:1 between the training and validation sets to avoid overfitting problems.

\textbf{The testing dataset} includes four real HAD datasets, namely San Diego (SD), Los Angeles (LA), Texas Coast (TC), and Hydice. The first three datasets are collected by the Airborne Visible/Infrared Imaging Spectrometer (AVIRIS) sensor and named with their capture places, while the last dataset shares the name with its captured sensor, i.e., Hyperspectral Digital Imagery Collection Experiment (HYDICE) airborne sensor. It is worth noting that the anomalies in these datasets vary in size and shape, representing different objects. Even if some datasets may have the same type of anomalies, there will still be differences in detection due to variations in their spatial resolution, capture time and scenes. The detailed information about these datasets can be found in Table \ref{dataset}.

\begin{table*}[t]
\centering
\caption{Detailed Information for HAD Datasets}
\renewcommand{\arraystretch}{1.1}
\setlength\tabcolsep{0.4mm}{
\begin{tabular}{cccccc}
\hline
Datasets  & Size                            & Anomaly size & Anomaly type & Spatial resolution & Date      \\ \hline
SD        & 100$\times$100$\times$189       & 134 pixels   & airplanes    & 3.5 m              & unknown   \\
LA        & 100$\times$100$\times$205       & 87 pixels    & airplanes    & 7.1 m              & 11/9/2011 \\
TC        & 100$\times$100$\times$198       & 155 pixels   & houses       & 17.2 m             & 8/29/2010 \\
Hydice    & 100$\times$80$\times$162        & 19 pixels    & vehicles     & 1 m                & unknown   \\ \hline
\end{tabular}
}\label{dataset}
\end{table*}  \par

\subsection{Experimental Setup}
\label{subsec42}
\textbf{Evaluation Metrics:} The 3D receiver operating characteristic (ROC) \citep{3DROC} curve is used to qualitatively compare the performance of different HAD methods. Specifically, its unfolded form, the 2D-ROC curves of $\left( {{P_d},{\rm{ }}{P_f}} \right)$, $\left( {{P_d},{\rm{ }}\tau } \right)$, and $\left( {{P_f},{\rm{ }}\tau } \right)$, can intuitively reflect the overall detection effectiveness, the target detection accuracy, and the background suppressibility. Additionally, the area under these 2D-ROC curves (AUC) \citep{AUC} is also used for quantitative analysis. Notably, two comprehensive evaluation indicators, ${\rm{AU}}{{\rm{C}}_{OA}}$ and ${\rm{AU}}{{\rm{C}}_{SNPR}}$ \citep{3DROC}, are also involved in the results analysis, which are defined based on the above three AUC indicators:
\begin{align}
{\rm{AU}}{{\rm{C}}_{OA}} = {\rm{AU}}{{\rm{C}}_{\left( {{P_d},{\rm{ }}{P_f}} \right)}} + {\rm{AU}}{{\rm{C}}_{\left( {{P_d},{\rm{ }}\tau } \right)}} - {\rm{AU}}{{\rm{C}}_{\left( {{P_f},{\rm{ }}\tau } \right)}}
\label{AUC1}
\end{align}
\begin{align}
{\rm{AU}}{{\rm{C}}_{SNPR}} = \frac{{{\rm{AU}}{{\rm{C}}_{\left( {{P_d},{\rm{ }}\tau } \right)}}}}{{{\rm{AU}}{{\rm{C}}_{\left( {{P_f},{\rm{ }}\tau } \right)}}}}
\label{AUC2}
\end{align}

\begin{figure}[t]
\centering
\includegraphics{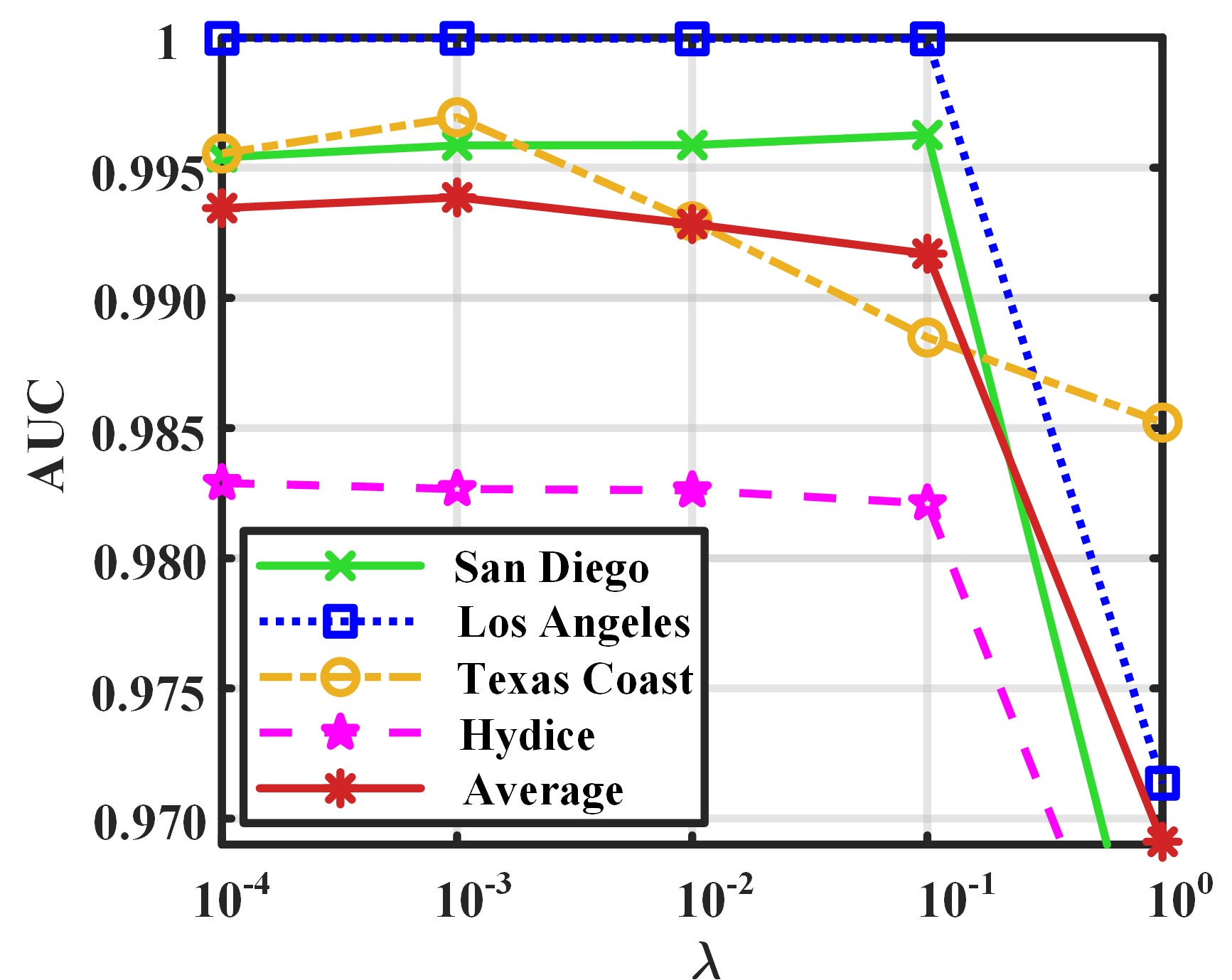}
\caption{Analysis of parameters $\lambda $ over experimental datasets}
\label{paras}
\end{figure}

\textbf{Parameter Settings:} Fig. \ref{paras} illustrates the variation of ${\rm{AU}}{{\rm{C}}_{\left( {{P_d},{\rm{ }}{P_f}} \right)}}$ over four experimental datasets with respect to $\lambda $. It can be observed that when $\lambda $ is greater than $10^{-2}$, the detection performance of four datasets drops significantly. Conversely, the detection performance shows little difference when $\lambda $ takes as $10^{-4}$, $10^{-3}$, or $10^{-2}$. Based on the highest average ${\rm{AU}}{{\rm{C}}_{\left( {{P_d},{\rm{ }}{P_f}} \right)}}$ over four datasets, we set $\lambda  = {10^{ - 3}}$.

\textbf{Comparison Methods:} Nine HAD methods are used for comparison, including LRASR \citep{LRASR}, graph and total variation regularized LRR model (GTVLRR) \citep{GTVLRR}, Fractional Fourier Entropy-based RX detector (FrFE) \citep{FrFE}, subspace selection-based discriminative forest model (SSDF) \citep{SSDF}, as well as three LSDM-based methods: the first decomposes the original HSI into third-order tensors called prior and tensor approximation model (PTA) \citep{PTA}, while the other two are LSDM-MoG \citep{MOG} and OSP-GoDec \citep{OSP}. Additionally, we use the DeCNN regularized anomaly detection method (DeCNN-AD) \citep{PAP} and autonomous HAD network (Auto-AD) \citep{Auto_AD} to showcase the advantages of SAP compared to the DNN-based method.

\begin{table*}[t]
\centering
\caption{Effectiveness Analysis of Self-Supervised Prior and Double-Purified Strategy on Experimental Datasets.}
\label{SSP_table} 
\renewcommand{\arraystretch}{1.26}
\setlength\tabcolsep{1mm}{
\begin{tabular}{c|ccc|ccc|ccc}
\hline
\multirow{2}{*}{Datasets} & \multicolumn{3}{c|}{AUC Values with SSP} & \multicolumn{3}{c|}{AUC Values with $L_{2,1}$} & \multicolumn{3}{c}{AUC Values with LREN} \\ \cline{2-10} 
                          & ${\left( {{P_d},{\rm{ }}{P_f}} \right)}$           & ${\left( {{P_d},{\rm{ }}\tau } \right)}$             & ${\left( {{P_f},{\rm{ }}\tau } \right)}$            & ${\left( {{P_d},{\rm{ }}{P_f}} \right)}$           & ${\left( {{P_d},{\rm{ }}\tau } \right)}$            & ${\left( {{P_f},{\rm{ }}\tau } \right)}$          & ${\left( {{P_d},{\rm{ }}{P_f}} \right)}$           & ${\left( {{P_d},{\rm{ }}\tau } \right)}$             & ${\left( {{P_f},{\rm{ }}\tau } \right)}$           \\ \hline
SD                 & 0.99586      & 0.50340     & 0.01431     & 0.99366      & 0.47715     & 0.05688     & 0.86998      & 0.05195     & 0.05153     \\
LA               & 0.98266      & 0.24584     & 0.01415     & 0.94298      & 0.22553     & 0.02409     & 0.94535      & 0.42996     & 0.06150     \\
TC               & 0.99612      & 0.38859     & 0.01367     & 0.99579      & 0.37729     & 0.03838     & 0.99639       & 0.66889     & 0.04750     \\
Hydice                    & 0.99997      & 0.66204     & 0.01587     & 0.99995      & 0.65591     & 0.13450     & 0.88850      & 0.21849     & 0.04001     \\
Average                   & 0.99365      & 0.42766     & 0.01817     & 0.98310      & 0.43397     & 0.06346     & 0.92506      & 0.34232     & 0.05014     \\ \hline
\end{tabular}
}
\end{table*}

\begin{figure}[htp]
\centering
\includegraphics[width=\textwidth]{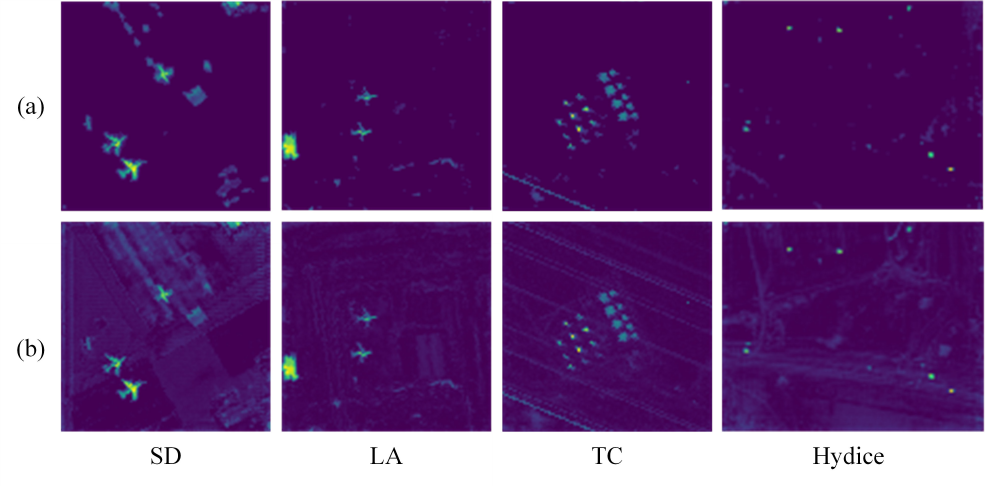}
\caption{Visual detection results with different anomaly priors on four HAD datasets. (a) With self-supervised prior. (b) With $\ell_{2,1}$-norm.}
\label{SSP_L21}
\end{figure}  \par

\subsection{Effectiveness Assessment}
\label{subsec43}
To better demonstrate the effectiveness of the main innovations in SAP (i.e., the self-supervised prior (SSP) and dual-purified strategy), we conduct quantitative analysis through ablation experiments in this section. \par

Table \ref{SSP_table} presents the AUC values of using $\ell_{2,1}$-norm to solve the objective function (\ref{objective-function}). In other words, these detection results are obtained by changing the updating ${\bf{A}}$ function (\ref{E}) to
\begin{align}
\mathop {\min }\limits_{\bf{A}} {\left\| {\bf{A}} \right\|_{2,1}} + \frac{1}{2}\left\| {\varphi \left( {\bf{H}} \right) - {\bf{DE}} - {\bf{A}}} \right\|_{_F}^2
\end{align}
Obviously, the AUC values obtained by SSP outperform those obtained with $\ell_{2,1}$-norm. In particular, the ${\rm{AU}}{{\rm{C}}_{\left( {{P_f},{\rm{ }}\tau } \right)}}$ values for the $\ell_{2,1}$-norm are more than twice as high as those for SSP across most datasets. This discrepancy suggests that without manually tuning the parameters, using the $\ell_{2,1}$-norm for anomaly detection leads to higher false alarms.
The visual results in Fig. \ref{SSP_L21} corroborate this finding, where more background information is preserved in the SD dataset's $\ell_{2,1}$-norm detection map, while the strip noise is detected along with the anomalies in the TC dataset's $\ell_{2,1}$-norm detection map. These observations further supports that the $\ell_{2,1}$-norm is not sufficiently appropriate to describe hyperspectral anomalies since it can also represent the noise in HSI \citep{PABDC}.

As for the dual-purified strategy, the ablation experiment is designed to use the original LREN \citep{AAAI} to construct the dictionary and latent HSI rather than the modified dictionary we proposed in Section \ref{subsec33}. As shown in Table \ref{SSP_table}, although the ${\rm{AU}}{{\rm{C}}_{\left( {{P_d},{\rm{ }}\tau } \right)}}$ values with LREN are higher on the LA and TC datasets, the detection performance are significantly improved with modified dictionary on the other two datasets. These improvements can be attributed to the fact that the modified dictionary constructed with dual-purified strategy has enough atoms to provide a finer description of the complex background, which is difficult for LREN. 
Therefore, the distinction between background and anomaly in LREN's results is not clear for HSIs with complex background, like the SD and Hydice datasets, resulting in lower detection accuracy for the anomaly targets.

\begin{figure*}[htp]
\centering
\includegraphics[width=\textwidth]{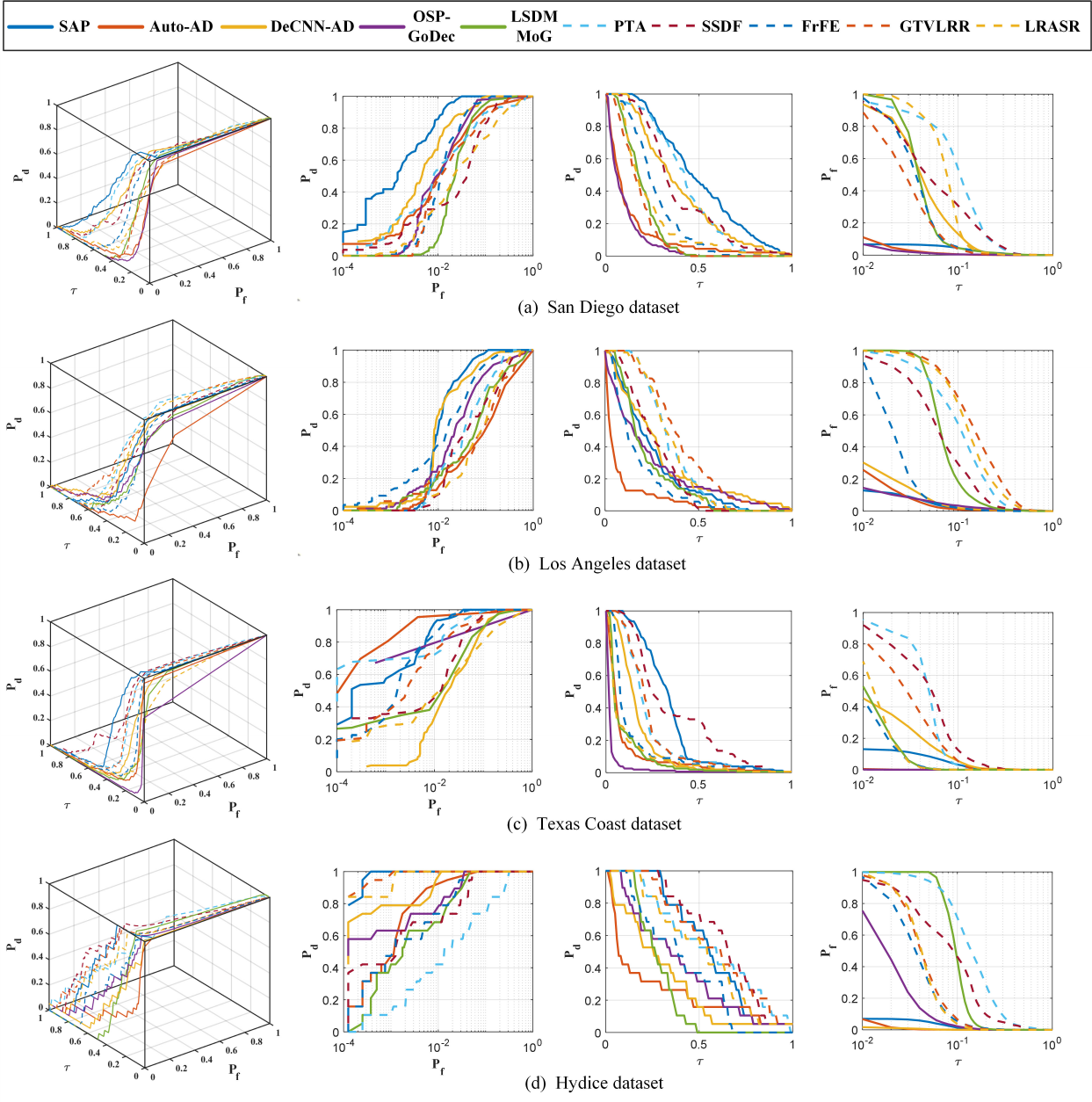}
\caption{ROC curves of all comparing HAD methods on the four experimental datasets.}
\label{ROCfig}
\vspace{-0.1in}
\end{figure*}  \par

\subsection{Detection Performance}
\label{subsec44}
As shown in Fig. \ref{ROCfig}, the SAP's $\left( {{P_d},{\rm{ }}{P_f}} \right)$ and $\left( {{P_d},{\rm{ }}\tau } \right)$ curves are positioned above all the other compared methods in the SD dataset, meaning that SAP is the best among these ten HAD methods in overall detection performance and target detection accuracy. Similarly, the $\left( {{P_d},{\rm{ }}{P_f}} \right)$ curve of SAP also occupies the upper left corner in the Hydice dataset. Nonetheless, the intersecting ROC curves bring resistance to comparative analysis. For example, we cannot explicitly determine the ranking of these ten methods in terms of $\left( {{P_d},{\rm{ }}\tau } \right)$ on the LA and Hydice datasets from Fig. \ref{ROCfig}. Thus, Tables \ref{San DiegoAUC}-\ref{HYDICEAUC} provide the AUC values of five indicators for all comparing methods, in which the top two results are emphasized with boldface and underline. \par

\begin{figure*}[t]
\centering
\includegraphics[width=\textwidth]{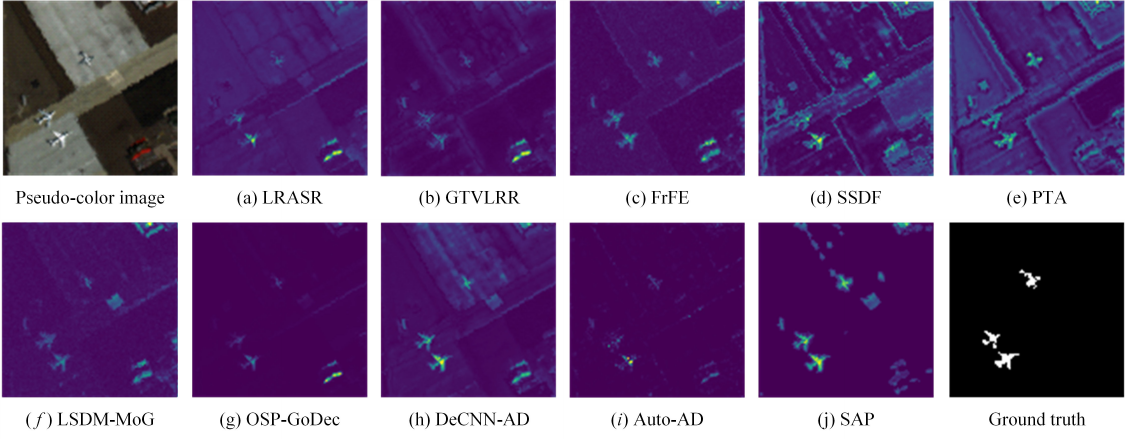}
\caption{Detection maps of the compared methods for SD dataset.}
\label{fig_SD}
\end{figure*}

According to Table \ref{San DiegoAUC}, SAP achieves the best performance in four out of five AUC indicators on the SD dataset. Even though its ${\rm{AU}}{{\rm{C}}_{\left( {{P_f},{\rm{ }}\tau } \right)}}$ falls short of the best, it still ranks third. The lowest and second lowest ${\rm{AU}}{{\rm{C}}_{\left( {{P_f},{\rm{ }}\tau } \right)}}$ values are obtained by Auto-AD and OSP-GoDec, but their target detection accuracy is too low to see the anomaly target clearly (see Fig. \ref{fig_SD}). The same situation also occurs in GTVLRR and LSDM-MoG. Even worse, these two methods fail to completely separate the background information, causing the airplanes to blend with the background. Despite PTA, DeCNN-AD, and SSDF possessing high ${\rm{AU}}{{\rm{C}}_{\left( {{P_d},{\rm{ }}\tau } \right)}}$ values, they retain a substantial amount of background information, lacking competitive advantage to our proposed SAP.

  \par

\begin{figure*}[t]
\centering
\includegraphics[width=\textwidth]{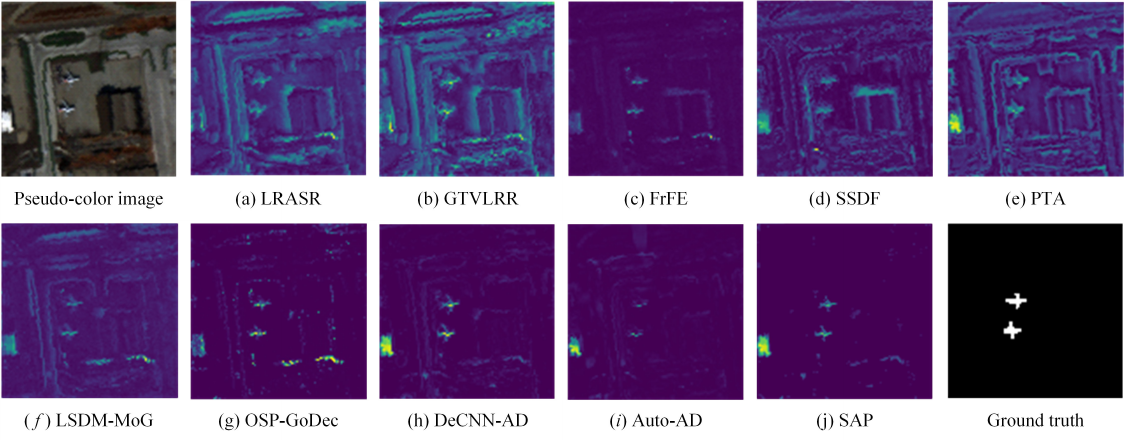}
\caption{Detection maps of the compared methods for LA dataset.}
\label{fig_LA}
\end{figure*}  \par

\begin{table}[htp]
\centering
\caption{AUC Values of the Compared Methods on the SD Dataset.}
\label{San DiegoAUC} 
\renewcommand{\arraystretch}{1.42}
\setlength\tabcolsep{0.26mm}{
\begin{tabular}{c|ccccc}
\hline
Method    & ${\rm{AU}}{{\rm{C}}_{\left( {{P_d},{\rm{ }}{P_f}} \right)}}$ & ${\rm{AU}}{{\rm{C}}_{\left( {{P_d},{\rm{ }}\tau } \right)}}$ & ${\rm{AU}}{{\rm{C}}_{\left( {{P_f},{\rm{ }}\tau } \right)}}$ & ${\rm{AU}}{{\rm{C}}_{OA}}$           & ${\rm{AU}}{{\rm{C}}_{SNPR}}$          \\ \hline
LRASR     & 0.91458                                                      & 0.22468                                                      & 0.08012                                                      & 1.05914                              & 2.80429                               \\
GTVLRR    & 0.95472                                                      & 0.17078                                                      & 0.03665                                                      & 1.08885                              & 4.65975                               \\ \hline
FrFE      & 0.98150                                                      & 0.26401                                                      & 0.04528                                                      & 1.20023                              & 5.83061                               \\
SSDF      & 0.94365                                                      & 0.39135                                                      & 0.05954                                                      & 1.27546                              & 6.57289                               \\
PTA       & 0.93919                                                      & \underline{0.42214}                         & 0.11980                                                      & 1.24153                              & 3.52371                               \\
LSDM-MoG  & 0.96464                                                      & 0.18945                                                      & 0.04875                                                      & 1.10534                              & 3.88615                               \\ \hline
OSP-GoDec & 0.98128                                                      & 0.09325                                                      &  \underline{0.00656}  & 1.06797                              & 14.2149                               \\
DeCNN-AD  & \underline{0.98765}                         & 0.39801                                                      & 0.06875                                                      & \underline{1.31691} & 5.78924                               \\
Auto-AD   & 0.97093                                                      & 0.13276                                                      & \textbf{0.00578}                            & 1.09791                              & \underline{22.9689} \\
SAP       & \textbf{0.99586}                            & \textbf{0.50340}                            & 0.01431                                                      & \textbf{1.48495}    & \textbf{35.1782}     \\ \hline
\end{tabular}
}
\vspace{-0.1in}
\end{table}  \par

\begin{table}[htp]
\centering
\caption{AUC Values of the Compared Methods on the LA Dataset.}
\label{Los AngelesAUC} 
\renewcommand{\arraystretch}{1.42}
\setlength\tabcolsep{0.26mm}{
\begin{tabular}{c|ccccc}
\hline
Method    & ${\rm{AU}}{{\rm{C}}_{\left( {{P_d},{\rm{ }}{P_f}} \right)}}$ & ${\rm{AU}}{{\rm{C}}_{\left( {{P_d},{\rm{ }}\tau } \right)}}$ & ${\rm{AU}}{{\rm{C}}_{\left( {{P_f},{\rm{ }}\tau } \right)}}$ & ${\rm{AU}}{{\rm{C}}_{OA}}$           & ${\rm{AU}}{{\rm{C}}_{SNPR}}$         \\ \hline
LRASR     & 0.88737                                                      & 0.34209                                                      & 0.15602                                                      & 1.07344                              & 2.19260                              \\
GTVLRR    & 0.85911                                                      & \textbf{0.40634}                            & 0.17769                                                      & 1.08776                              & 2.28679                              \\ \hline
FrFE      & 0.96557                                                      & 0.17511                                                      & 0.02661                                                      & 1.11407                              & 6.58061                              \\
SSDF      & 0.89351                                                      & 0.25862                                                      & 0.08714                                                      & 1.06499                              & 2.96787                              \\
PTA       & 0.93226                                                      & \underline{0.34627}                         & 0.12471                                                      & 1.15382                              & 2.77660                              \\
LSDM-MoG  & 0.87545                                                      & 0.21573                                                      & 0.08134                                                      & 1.00984                              & 2.65220                              \\ \hline
OSP-GoDec & 0.93659                                                      & 0.23635                                                      & 0.01531                                                      & 1.15763                              & 15.4376                              \\
DeCNN-AD  & \underline{0.97403}                         & 0.29696                                                      & 0.01756                                                      & \textbf{1.25343}    & \underline{16.9112} \\
Auto-AD   & 0.84432                                                      & 0.07897                                                      & \textbf{0.01298}                            & 0.91031                              & 6.08398                              \\
SAP       & \textbf{0.98266}                            & 0.24584                                                      & \underline{0.01415}                         & \underline{1.21435} & \textbf{17.3739}    \\ \hline
\end{tabular}
}
\vspace{-0.1in}
\end{table}  \par

Fig. \ref{fig_LA} displays the LA dataset's detection maps over ten HAD methods, where half of the methods are incapable of distinguishing the airplanes (i.e., anomaly targets) from the edge part of the background scene, such as the outline of the taxiway. This is because both the anomalies and the edge parts belong to the high-frequency component of the image \citep{DFAE} and exhibit similarity in data distribution. Thus, the overall detection performance of GTVLRR, PTA, LRASR and SSDF is much lower than SAP in Table \ref{Los AngelesAUC}, even though their ${\rm{AU}}{{\rm{C}}_{\left( {{P_d},{\rm{ }}\tau } \right)}}$ values are higher than SAP. Similarly, although Auto-AD obtains a lower ${\rm{AU}}{{\rm{C}}_{\left( {{P_f},{\rm{ }}\tau } \right)}}$ than SAP, the airplanes detected in SAP exhibit a more complete morphology than those in Auto-AD.

\begin{figure*}[htp]
\centering
\includegraphics[width=\textwidth]{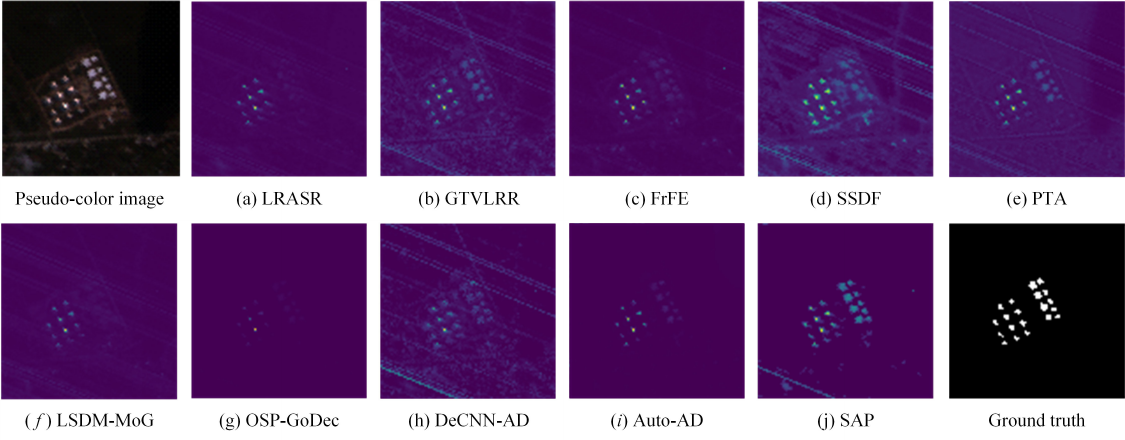}
\caption{Detection maps of the compared methods for TC dataset.}
\label{fig_urban}
\end{figure*}  \par

\begin{figure*}[htp]
\centering
\includegraphics[width=\textwidth]{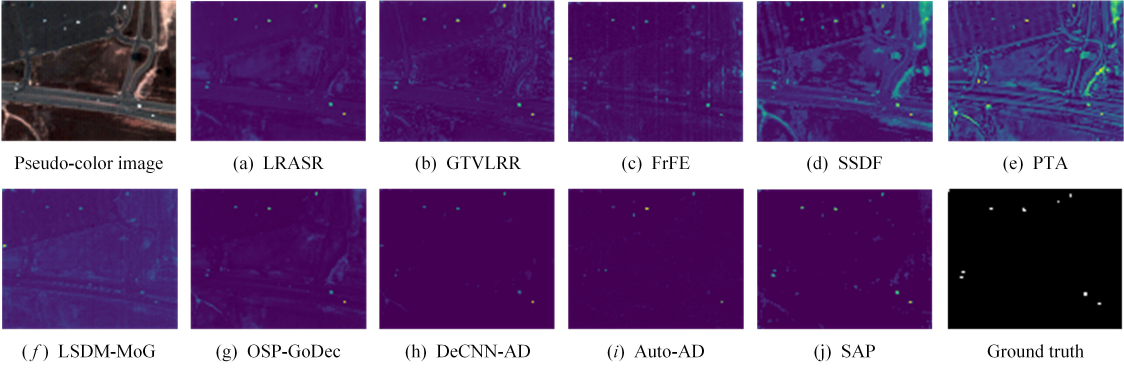}
\caption{Detection maps of the compared methods for Hydice dataset.}
\label{fig_hydice}
\end{figure*}  \par

\begin{table}[t]
\centering
\caption{AUC Values of the Compared Methods on the TC Dataset.}
\label{Texas CoastAUC} 
\renewcommand{\arraystretch}{1.42}
\setlength\tabcolsep{0.26mm}{
\begin{tabular}{c|ccccc}
\hline
Method    & ${\rm{AU}}{{\rm{C}}_{\left( {{P_d},{\rm{ }}{P_f}} \right)}}$ & ${\rm{AU}}{{\rm{C}}_{\left( {{P_d},{\rm{ }}\tau } \right)}}$ & ${\rm{AU}}{{\rm{C}}_{\left( {{P_f},{\rm{ }}\tau } \right)}}$                        & ${\rm{AU}}{{\rm{C}}_{OA}}$           & ${\rm{AU}}{{\rm{C}}_{SNPR}}$         \\ \hline
LRASR     & 0.93581                                                      & 0.12244                                                      & 0.01461                                                                             & 1.04364                              & 8.38056                              \\
GTVLRR    & 0.98489                                                      & 0.27192                                                      & 0.04044                                                                             & 1.21637                              & 6.72404                              \\ \hline
FrFE      & 0.99570                                                      & 0.15235                                                      & 0.01225                                                                             & 1.13580                              & 12.4367                              \\
SSDF      & 0.97804                                                      & \underline{0.35911}                                                      & 0.05954                                                                             & 1.27761                              & 6.03141                              \\
PTA       & 0.99156                                                      & 0.34699                                                      & 0.04126                                                                             & \underline{1.29729} & 8.40984                              \\
LSDM-MoG   & 0.96731                                                      & 0.10675                                                      & 0.01368                                                                             & 1.06038                              & 7.80336                              \\ \hline
OSP-GoDec & \underline{0.99850}  & 0.04246                                                      & \textbf{0.00046}                                                   & 1.04050                              & \textbf{92.3043}    \\
DeCNN-AD  & 0.95697                                                      & 0.17879                                                      & 0.02473                                                                             & 1.11103                              & 7.22968                              \\
Auto-AD   & \textbf{0.99882}     & 0.23388  & \underline{0.00286} & 1.22984                              & \underline{81.7762} \\
SAP       & 0.99612                                                      & \textbf{0.38859}                            & 0.01167                                                                             & \textbf{1.37304}    & 33.2982                              \\ \hline
\end{tabular}
}
\vspace{-0.1in}
\end{table}  \par

Taking Fig. \ref{fig_urban}, GTVLRR, SSDF, PTA, and DeCNN-AD fail to eliminate the inherent stripe noise in the Texas Coast dataset, affecting their overall detection performance. Although LRASR and LSDM-MoG eliminate most of the stripe noise, their detection results are less evident, especially for the anomaly targets in the right half, which are almost undetected, resulting in their low ${\rm{AU}}{{\rm{C}}_{\left( {{P_d},{\rm{ }}{P_f}} \right)}}$ values (see Table \ref{Texas CoastAUC}). In comparison, the detection results of OSP-GoDec and Auto-AD are completely unaffected by noise. However, the visual results in OSP-GoDec also loses the anomaly targets of interest. This is disadvantageous for practical applications because anomaly targets could be disaster warning signals and missing such anomaly targets would lead to severe consequences. Overall, Auto-AD achieves the best detection performance, followed by our proposed SAP. Compared to Auto-AD, SAP has a higher false alarm rate. How to reduce these false alarms and improve the background suppression of SAP is our future research direction.

\begin{table}[htp]
\centering
\caption{AUC Values of the Compared Methods on the Hydice Dataset.}
\label{HYDICEAUC} 
\renewcommand{\arraystretch}{1.42}
\setlength\tabcolsep{0.26mm}{
\begin{tabular}{c|ccccc}
\hline
Method    & ${\rm{AU}}{{\rm{C}}_{\left( {{P_d},{\rm{ }}{P_f}} \right)}}$ & ${\rm{AU}}{{\rm{C}}_{\left( {{P_d},{\rm{ }}\tau } \right)}}$ & ${\rm{AU}}{{\rm{C}}_{\left( {{P_f},{\rm{ }}\tau } \right)}}$                        & ${\rm{AU}}{{\rm{C}}_{OA}}$           & ${\rm{AU}}{{\rm{C}}_{SNPR}}$         \\ \hline
LRASR     & 0.99983                                                      & 0.61369                                                      & 0.09345                                                                             & 1.52007                              & 6.56704                              \\
GTVLRR    & \underline{0.99991}                         & 0.64166                                                      & 0.12088                                                                             & 1.52069                              & 5.30824                              \\ \hline
FrFE      & 0.99258                                                      & 0.42088                                                      & 0.06202                                                                             & 1.35144                              & 6.78620                              \\
SSDF      & 0.98864                                                      & \underline{0.66191}                         & 0.10914                                                                             & \underline{1.54141} & 6.06478                              \\
PTA       & 0.93504                                                      & 0.56367                                                      & 0.15797                                                                             & 1.34074                              & 3.56821                              \\
LSDMMoG   & 0.99170                                                      & 0.27783                                                      & 0.08137                                                                             & 1.18816                              & 3.41440                              \\ \hline
OSP-GoDec & 0.99407                                                      & 0.47051                                                      & 0.04312                                                                             & 1.42146                              & 10.9116                              \\
DeCNN-AD  & 0.99819                                                      & 0.41522                                                      & \textbf{0.00093}                                                   & 1.41248                              & \textbf{446.473}    \\
Auto-AD   & 0.99746                               & 0.28281                               & \underline{0.00771} & 1.27256                              & 36.6809                              \\
SAP       & \textbf{0.99997}                            & \textbf{0.67116}                            & 0.01587                                                                             & \textbf{1.65526}    & \underline{42.2911} \\ \hline
\end{tabular}
}
\vspace{-0.1in}
\end{table}  \par

As for the Hydice dataset, the four AUC indicators of SAP are either first or second among all compared methods (see Table \ref{HYDICEAUC}), only the ${\rm{AU}}{{\rm{C}}_{\left( {{P_f},{\rm{ }}\tau } \right)}}$ value falls behind DeCNN-AD and Auto-AD. However, DeCNN-AD's low ${\rm{AU}}{{\rm{C}}_{\left( {{P_f},{\rm{ }}\tau } \right)}}$ stems from its manually optimized parameters, which are not applicable to other datasets. When detecting different HSIs, DeCNN-AD requires a new search for the best parameters, making it less flexible and generalizable to SAP. Additionally, it is evident that Auto-AD fails to detect the leftmost two anomaly targets from Fig. \ref{fig_hydice}, which also effect its overall detection performance. Thus, the proposed SAP remains highly competitive on the Hydice dataset.

\section{Conclusion}
\label{sec5}
In this paper, we propose an LRR model with self-supervised anomaly prior (SAP) for HAD. The purpose of our method is to eliminate the limitations of handcrafted anomaly prior. For achieving this, multiple aspects are considered. Firstly, given the problem of lacking labels and training samples, we employ self-supervised learning to train the network, where the pretext task is designed as a classification task to distinguish the original HSI and the pseudo-anomaly HSI generated from the original HSI. Secondly, considering the sparsity and spatial characteristics of hyperspectral anomalies, the pseudo-anomaly is customized as a prism with arbitrary polygon bases and arbitrary spectral bands, which is also cropped and pasted at any position in both spatial and spectral domains to generalize various anomalies. After completing the pretext task, the body part of the well-trained network serves as the anomaly prior to solve the LRR model by the plug-and-play strategy. Additionally, we propose a dual-purified strategy to construct an enriched background dictionary without anomaly contamination, thereby assisting in separating anomalies from complex backgrounds by providing a finer background representation. Extensive experiments on real datasets demonstrate that the proposed SAP achieves state-of-the-art results while retaining the theoretical explanation of the mathematical optimization model.

\section*{Acknowledgments}
  This work was supported in part by the National Natural Science Foundation of China under Grant 62121001 and Grant U22B2014.

 \bibliographystyle{elsarticle-harv} 
 \bibliography{egref}






\end{document}